%% file: main.tex
\documentclass[10pt]{article}

\usepackage[a4paper,margin=1in]{geometry}
\usepackage[T1]{fontenc}
\usepackage[utf8]{inputenc}
\usepackage[english]{babel}
\usepackage{hyperref}
\usepackage{amsmath,amssymb}
\usepackage{booktabs}
\usepackage{enumitem}
\usepackage{xcolor}
\usepackage{graphicx}
\usepackage{microtype}
\usepackage{siunitx}
\usepackage{float}
\usepackage{placeins}
\usepackage{subcaption}
\usepackage{orcidlink}
\hypersetup{
  colorlinks=true,
  linkcolor=[rgb]{0.0,0.2,0.6},
  urlcolor=[rgb]{0.0,0.2,0.6},
  citecolor=[rgb]{0.0,0.2,0.6}
}

\title{\textbf{Edge-aware baselines for \texttt{ogbn-proteins} in PyTorch Geometric}\\
\large Species-wise normalization, post-hoc calibration, and cost--accuracy trade-offs}

\author{
  \href{https://orcid.org/0009-0003-5238-7251}{Aleksandar Stankovi\'c}\orcidlink{0009-0003-5238-7251}\thanks{Faculty of Technical Sciences, University of Novi Sad. \texttt{stankovic.sv25.2022@uns.ac.rs}}
  \and
  Dejan Lisica\thanks{Faculty of Technical Sciences, University of Novi Sad. \texttt{lisica.sv49.2022@uns.ac.rs}}
}
\date{\today}

\begin{document}
\maketitle

\begin{abstract}
\noindent
We present reproducible, edge-aware baselines for \texttt{ogbn-proteins} in PyTorch Geometric (PyG). We study two system choices that dominate practice: (i) how 8-D edge evidence is aggregated into node inputs, and (ii) how edges are used inside message passing. Our strongest baseline is GraphSAGE with \emph{sum}-based edge$\to$node features. We compare LayerNorm (LN), BatchNorm (BN), and a species-aware Conditional LayerNorm (CLN), and report compute cost (time, VRAM, parameters) together with accuracy (ROC-AUC) and decision quality. In our primary experimental setup (hidden=512, 3 layers, 3 seeds), \emph{sum} consistently beats \emph{mean}/\emph{max}; BN attains the best AUC, while CLN matches the AUC frontier with better thresholded F1. Finally, post-hoc \emph{per-label} temperature scaling plus per-label thresholds substantially improves micro-F1 and ECE with negligible AUC change, and light label-correlation smoothing yields small additional gains. We release standardized artifacts and scripts used for all of the runs presented in the paper.
\end{abstract}

\section{Introduction}
Many real-world problems are graphs: nodes are entities and edges describe relations. In \texttt{ogbn-proteins}~\cite{ogb}, nodes are proteins and edges carry an 8-D vector of association evidence; each protein has 112 binary functional labels. The official metric is mean ROC-AUC across labels, with a \emph{species-wise} split: validation and test contain species unseen during training.

This work focuses on two practical design questions: \textbf{(i)} how to construct node inputs from incident 8-D edge features; and \textbf{(ii)} how to incorporate edge information inside message passing. We then compare normalization schemes (BN/LN/CLN) and report compute cost. Our strongest baseline is GraphSAGE, trained and reported under a standardized artifact format for reproducibility.

\section{Background}
\paragraph{Problem.} Given a directed graph $G=(V,E)$, an edge $(i,j)$ has $\mathbf{e}_{ij}\!\in\![0,1]^8$. A node $i$ has a multi-label target $\mathbf{y}_i\!\in\!\{0,1\}^{112}$. Models output per-label logits $\hat{\mathbf{z}}_i\!\in\!\mathbb{R}^{112}$.

\paragraph{Message passing.}
Modern GNNs iteratively update node states via neighbor aggregation~\cite{mpnn}:
\[
\mathbf{h}^{(\ell+1)}_i=\sigma\!\Big(\mathbf{W}_{\text{self}}\mathbf{h}^{(\ell)}_i+
\operatorname{AGG}_{j\in\mathcal{N}(i)} \alpha_{ij}\,\mathbf{W}\mathbf{h}^{(\ell)}_j + \mathbf{b}\Big).
\]

GraphSAGE~\cite{sage} uses (weighted) means; GIN~\cite{gin} uses sum-like updates with an MLP. In edge-network MPNNs~\cite{mpnn}, the edge weight $\alpha_{ij}$ is learned from $\mathbf{e}_{ij}$ \cite{li2016ggnn,bresson2017resgated}.

\paragraph{Normalization.} \textbf{BN}~\cite{bn} normalizes per batch; \textbf{LN}~\cite{ln} per node. \textbf{Conditional LN (CLN)} conditions LN’s affine parameters on auxiliary context (here: species), following conditional normalization ideas~\cite{devries2017modulating,huang2017adain,perez2018film}.

\section{Related Work}

\paragraph{GNNs with edge information.}
Message-passing neural networks (MPNNs) provide a general template for incorporating edge features via learned gates or edge-conditioned weights~\cite{mpnn}. Variants such as Gated Graph Neural Networks and Residual Gated Graph ConvNets explicitly modulate messages with edge-dependent gating~\cite{li2016ggnn,bresson2017resgated}. Inductive baselines like GraphSAGE~\cite{sage} aggregate neighbor states with (weighted) means, whereas GIN~\cite{gin} emphasizes sum-like updates. Our study is complementary: rather than proposing a new architecture, we quantify how simple, \emph{edge-aware} choices—(i) aggregating 8-D edge evidence into node inputs; (ii) scalarizing edge channels to weight messages—shift the accuracy–cost frontier on \texttt{ogbn-proteins}.

\noindent
Our baselines also sit alongside work that studies the expressive power of GNNs for multi-node prediction tasks such as link prediction~\cite{zhang2020linkpred} and identity-/position-aware message passing~\cite{you2019pgnn,you2021idgnn}.

\paragraph{Normalization and conditioning.}
Normalization is a core component in deep GNNs. BatchNorm \cite{bn} and LayerNorm \cite{ln} are standard, while conditional normalization methods modulate affine parameters using side information \cite{devries2017modulating,huang2017adain,perez2018film}. We adopt a lightweight Conditional LayerNorm (CLN) that conditions on species descriptors and test whether it improves cross-species generalization under the official split (mouse$\to$zebrafish).

\paragraph{Calibration and decision thresholds.}
Neural networks are often miscalibrated. Temperature scaling provides a simple, effective post-hoc fix \cite{calib}; broader post-hoc calibrators include Platt scaling \cite{platt1999}, isotonic regression for probabilistic outputs \cite{zadrozny2002}, Bayesian binning into quantiles \cite{naeini2015bbq}, and Dirichlet calibration \cite{kull2019dirichlet}. While most prior reports on \texttt{ogbn-proteins} focus on ROC-AUC, we pair calibration with \emph{per-label} thresholds and report ECE and Brier score \cite{brier1950}, showing large gains in micro-F1 at essentially unchanged AUC.

\paragraph{Exploiting label dependencies.}
Multi-label methods often leverage inter-label structure, e.g., Classifier Chains \cite{read2009cc} or GCNs on a learned label graph (ML-GCN) \cite{mlgcn2019}. To keep the baseline simple and leakage-free, we compute a label co-occurrence matrix from training labels only and apply a small logit-space smoothing; we treat full label-graph learning as future work.

\section{Dataset and setup}
We follow the OGB \texttt{ogbn-proteins} protocol~\cite{ogb}. Initial node features are built by aggregating incident 8-D edge features with \textbf{mean}, \textbf{sum}, or \textbf{max}. We train on training species, validate on an unseen species, and test on a different unseen species. Hardware: NVIDIA A800-SXM4-40GB.

\paragraph{Species split.} The validation split contains a single species (NCBI \textbf{10090}, mouse); the test split also contains a single species (\textbf{7955}, zebrafish). Our per-species plots (Sec.~\ref{sec:per-species}) therefore display one bar per split while comparing model variants.

\paragraph{Metrics.} Primary: mean ROC-AUC across 112 labels. Secondary: micro-F1 at a fixed 0.5 threshold.

\paragraph{Implementation.} PyTorch Geometric~\cite{pyg}. Seeds $\{1,2,3\}$; early stopping on validation AUC. Each run exports \texttt{args.json}, \texttt{metrics.json}, and \texttt{logits\_\{val,test\}.npz} with \texttt{node\_id}, \texttt{species\_id}, \texttt{logits[112]}, \texttt{labels[112]}.

\noindent\textbf{Repository.} \url{https://github.com/SV25-22/ECHO-Proteins}

\section{Methods}

\paragraph{Edge$\to$node feature construction.}
For each node $i$, we aggregate features of incident edges $\{\mathbf{e}_{ji}\}$ into a node input $\mathbf{x}_i$ using \textsc{mean}, \textsc{sum}, or \textsc{max}. This isolates the contribution of edge evidence to node representations.

\paragraph{MLP (no-graph) baseline.}
To isolate the effect of message passing, we train a 3-layer MLP directly on the edge$\to$node features $\mathbf{x}_i$ to predict 112 logits (BCEWithLogits). Each hidden layer uses BN/LN/CLN $+$ LeakyReLU $+$ dropout. Training mirrors the GNN setup. See Sec.~\ref{sec:mlp}.

\paragraph{GraphSAGE / GIN with scalarized edges.}
We reduce $\mathbf{e}_{ij}\in\mathbb{R}^8$ to a scalar $\alpha_{ij}\!\in\!\mathbb{R}_+$ (default: channel \textsc{sum}; we also test a learned 1-D scalarizer), and use $\alpha_{ij}$ to weight messages in SAGE/GIN. This retains a simple architecture while exploiting edge strength.

\paragraph{Add-on A: Normalization variants.}
We compare \textbf{LN}, \textbf{BN}, and \textbf{CLN} (LN whose affine parameters are predicted from a per-species descriptor; \texttt{cln\_mode=desc}). Unless stated otherwise, SAGE uses hidden=512 and 3 layers.

\paragraph{Add-on B: Calibration \& thresholds.}
We apply post-hoc temperature scaling to validation logits to correct over- or under-confidence \cite{platt1999,zadrozny2002,naeini2015bbq,kull2019dirichlet}. We consider two modes: a single global temperature parameter and per-label temperatures with L2 regularization toward the global value. Temperatures are optimized by minimizing the negative log-likelihood on the validation set. After calibration, we derive per-label thresholds by maximizing the F$_\beta$ score on validation probabilities, falling back to ROC-based thresholds in degenerate cases. At test time, we apply the learned calibration and thresholds without re-fitting. To quantify calibration quality we compute the Expected Calibration Error (ECE) and the Brier score, in addition to ROC-AUC and F1.

\paragraph{Add-on C: Label Correlation Analysis}

We construct a label–label co-occurrence matrix $P \in \mathbb{R}^{K\times K}$ ($K=112$) from the training set labels. Each entry $P_{jk}$ encodes how frequently label $j$ and $k$ appear together relative to the total occurrences of $j$, with rows normalized to sum to one. Predictions are then smoothed using:
\[
    z' = z + \lambda\, z P^\top
\]
where $z$ are the raw logits and $\lambda$ is a small smoothing coefficient tuned on the validation set. This formulation allows information from correlated labels to adjust logits before thresholding. All correlation matrices are computed only from training labels to avoid data leakage \cite{read2009cc,mlgcn2019}.

\pagebreak

\section{Results}
All means and standard deviations are over the three seeds $\{1,2,3\}$.

\subsection{Main baselines}
\label{sec:main}
\input{tables/main_gnn_baselines}

\paragraph{Observations.}
With SAGE (sum, h{=}512, L{=}3; 3 seeds), \textbf{BN} attains the top ROC-AUC; \textbf{CLN} matches the AUC frontier while retaining much stronger fixed-threshold micro-F1. \textbf{LN} trails slightly in AUC but remains competitive with \emph{sum} edge$\to$node inputs. A learned 1-D edge scalarizer is close but does not surpass the simple \emph{sum}.\textbf{GIN+BN} lags SAGE in AUC (0.761 vs.\ $\sim$0.79) but shows a higher fixed-threshold F1 than SAGE+BN (0.149 vs.\ 0.096), highlighting threshold/calibration sensitivity; overall, SAGE remains the better choice on this benchmark.

\subsection{MLP baselines}
\label{sec:mlp}
To isolate the value of message passing, we train MLPs on edge-aggregated node features using mean, sum, and max aggregation. We evaluate 33 different configurations varying aggregation method, normalization (BatchNorm, LayerNorm, none), and architecture depth (uniform, tapering, deep).

\begin{table}[H]
\centering
\small
\begin{tabular}{lccccc}
\toprule
\textbf{MLP Configuration} & \textbf{Val AUC} & \textbf{Test AUC} & \textbf{Test micro-F1@0.5} & \textbf{VRAM (MB)} & \textbf{Params (M)}\\
\midrule
sum\_deep\_none (best)     & $0.796$ & $\mathbf{0.743}$ & $0.145$ & 847 & 0.181\\
sum\_taper\_bn             & $0.799$ & $\mathbf{0.743}$ & $0.128$ & 1075 & 0.185\\
sum\_deep\_bn              & $0.802$ & $\mathbf{0.742}$ & $0.125$ & 1073 & 0.183\\
sum\_uniform\_256\_bn      & $0.795$ & $\mathbf{0.740}$ & $0.137 $ & 732 & 0.098\\
max\_uniform\_256\_cln\_desc & $0.741$ & $\mathbf{0.715}$ & $0.120$ & 1021 & 0.098\\
\midrule
mean\_uniform\_256\_none   & $0.636$ & $\mathbf{0.577}$ & $0.075$ & 603 & 0.097\\
sum\_taper\_none (worst)   & $0.447$ & $\mathbf{0.465}$ & $0.049$ & 848 & 0.183\\
\bottomrule
\end{tabular}
\caption{MLP baselines on \texttt{ogbn-proteins}. SUM aggregation consistently outperforms MEAN and MAX. BatchNorm provides optimal performance with SUM aggregation, while no normalization performs poorly. The best MLP achieves 0.743 test AUC, demonstrating that edge aggregation alone captures significant signal.}
\label{tab:mlp}
\end{table}

\paragraph{Key findings.}
\textbf{Aggregation:} SUM aggregation dominates top performers (all top 4 models), achieving 0.74+ test AUC vs. 0.69-0.72 for others. 
\textbf{Normalization:} BatchNorm shows best performance with SUM aggregation, while no normalization performs poorly, especially with SUM. 
\textbf{Architecture:} Uniform width (256, 256) and tapering (512→256→128) perform similarly well. 
\textbf{Calibration:} SUM + BatchNorm models achieve the best calibration
($T_\mathrm{global} \approx 0.95$--$1.00$), while LayerNorm models are
overconfident ($T_\mathrm{global} > 1.05$).

\subsection{Edge-to-node aggregation ablation}
\label{sec:aggr}
\input{tables/aggr_ablation}

\begin{figure}[H]
  \centering
  \begin{subfigure}[t]{0.48\linewidth}
    \centering
    \includegraphics[width=\linewidth]{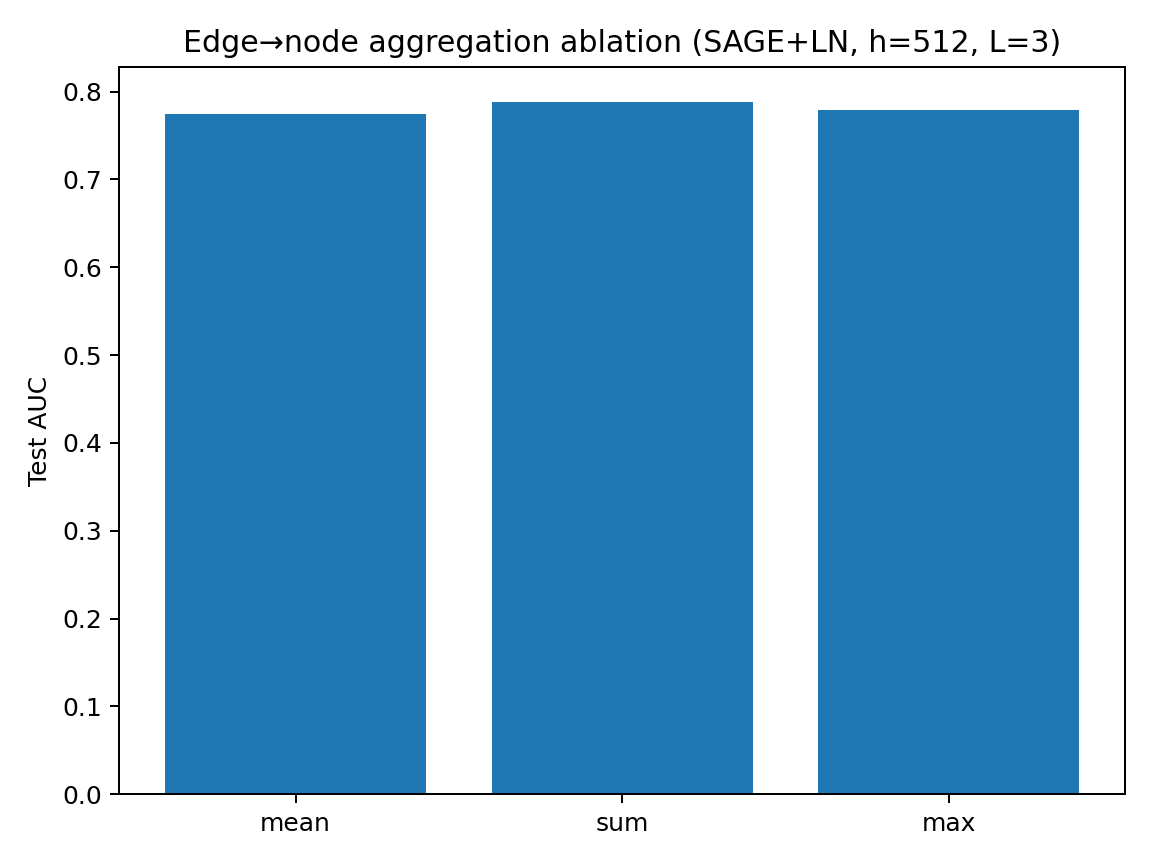}
    \caption{Test AUC}
  \end{subfigure}\hfill
  \begin{subfigure}[t]{0.48\linewidth}
    \centering
    \includegraphics[width=\linewidth]{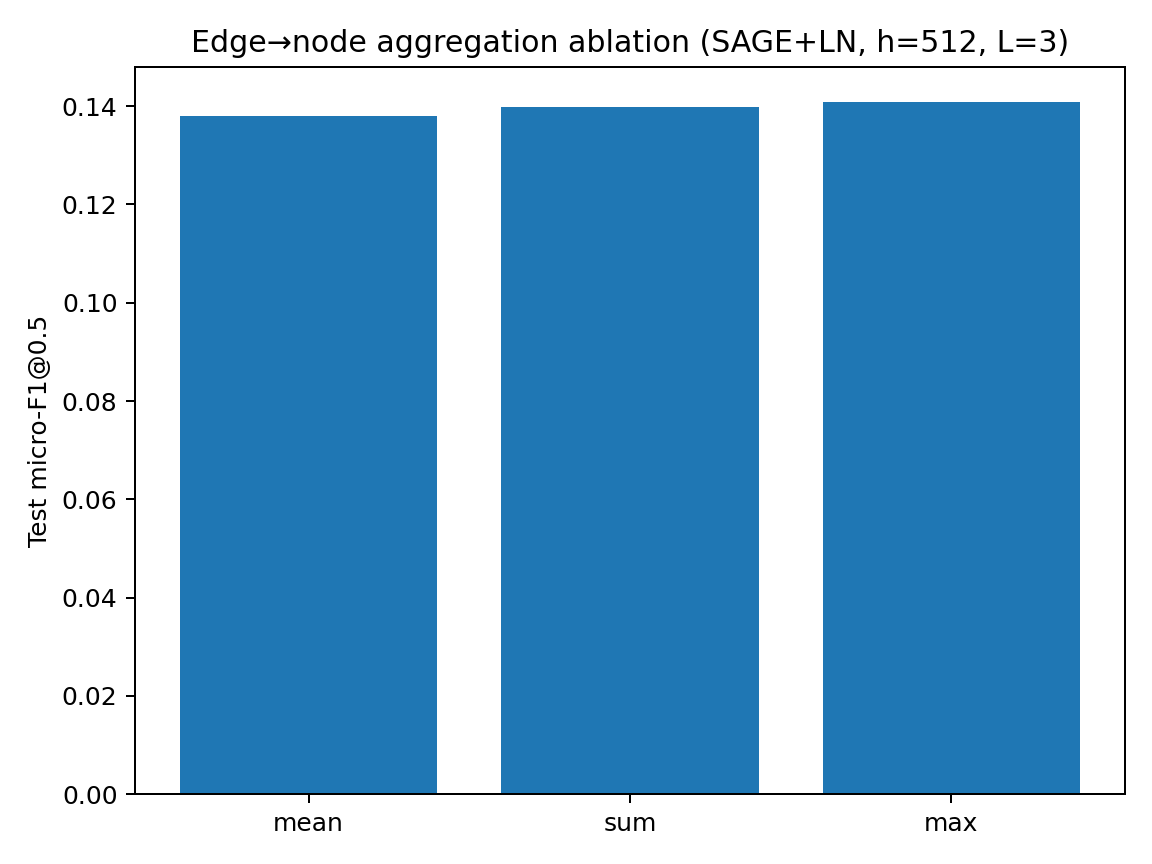}
    \caption{Test micro-F1@0.5}
  \end{subfigure}
  \caption{\textbf{Aggregation ablation} with SAGE+LN (3 seeds).}
  \label{fig:aggr-ablation}
\end{figure}

\paragraph{Takeaways.}
Replacing \textsc{mean} with \textsc{sum} for edge$\to$node features improves both AUC and micro-F1. \textsc{max} is competitive but slightly below \textsc{sum}. A simple learned 1-D scalarizer for edges is close to \textsc{sum} but does not consistently beat it.

\subsection{Per-species analysis}
\label{sec:per-species}
Because the benchmark has one validation species (mouse; 10090) and one test species (zebrafish; 7955), we compare normalization choices on those species using the new SAGE (hid=512, L=3) runs (seeds 1--3). BN and CLN are statistically tied on mouse within error bars; on zebrafish, CLN and BN remain close, with LN slightly behind. This supports CLN as a robust alternative to BN for cross-species generalization.

\begin{figure}[H]
  \centering
  \begin{subfigure}[t]{0.49\linewidth}
    \centering
    \includegraphics[width=\linewidth]{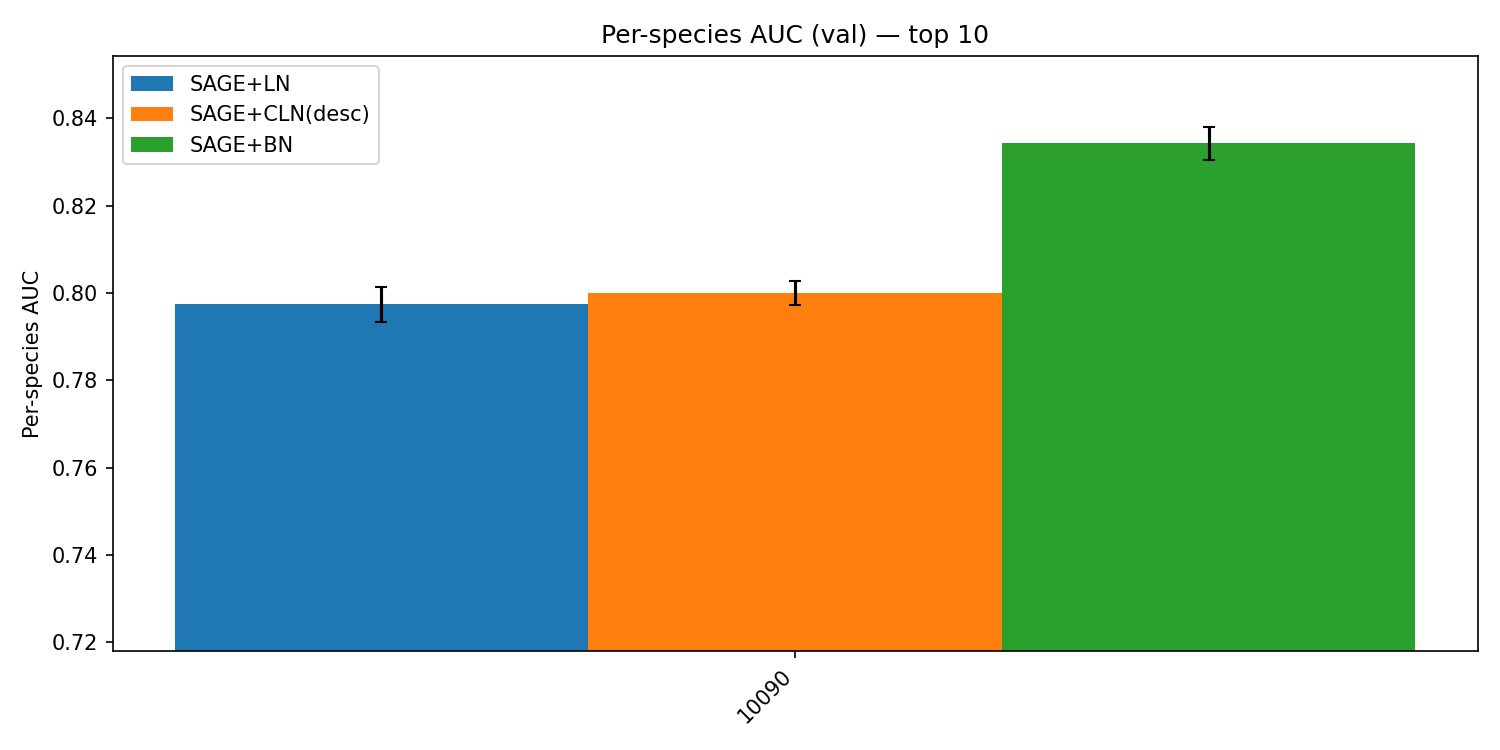}
    \caption{Validation species (10090)}
  \end{subfigure}\hfill
  \begin{subfigure}[t]{0.49\linewidth}
    \centering
    \includegraphics[width=\linewidth]{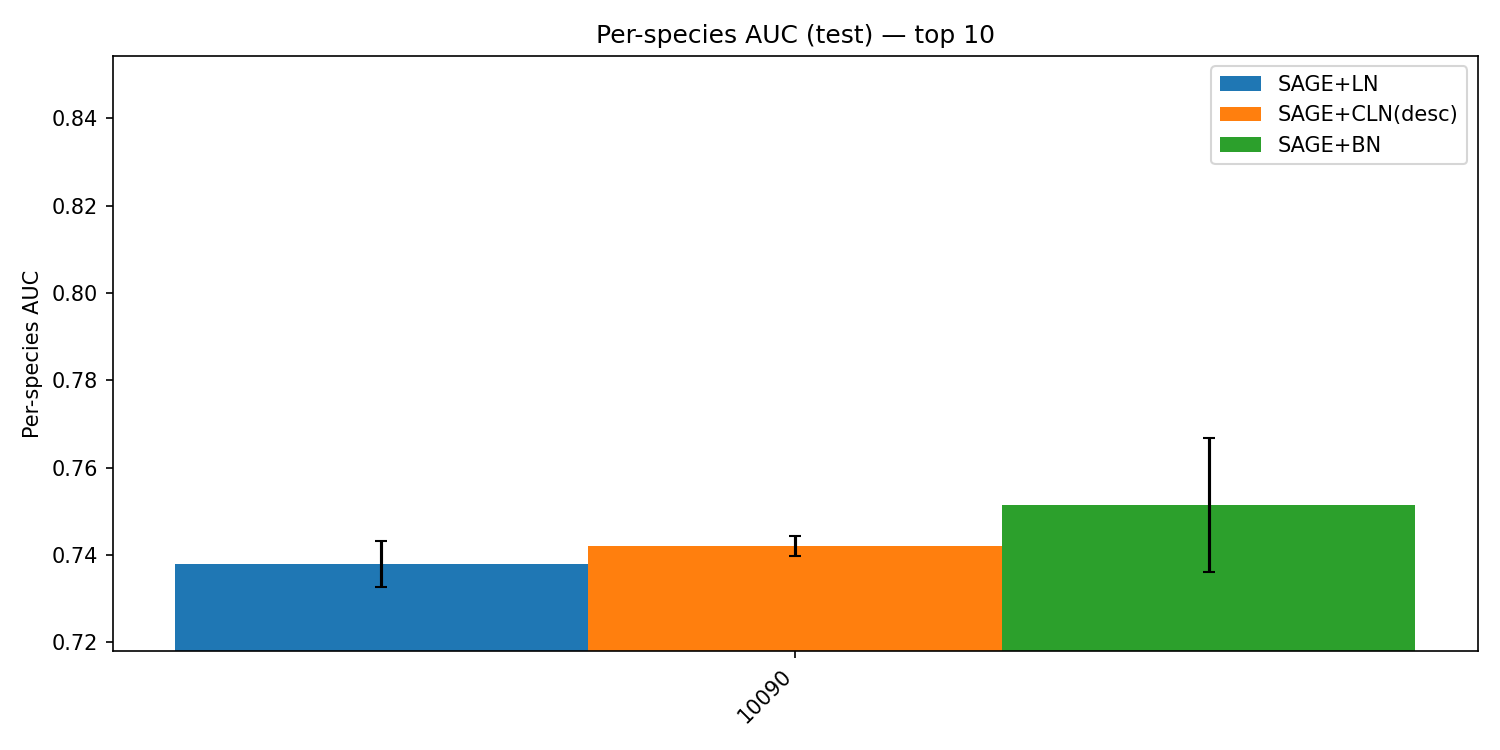}
    \caption{Test species (7955)}
  \end{subfigure}
  \caption{\textbf{Per-species AUC} (mean $\pm$ s.d., 3 seeds) for SAGE (hid=512, L=3) with LN/BN/CLN.}
  \label{fig:per-species}
\end{figure}

\subsection{AUC vs.\ cost}
\label{sec:cost}
\begin{figure}[H]
  \centering
  \includegraphics[width=.72\linewidth]{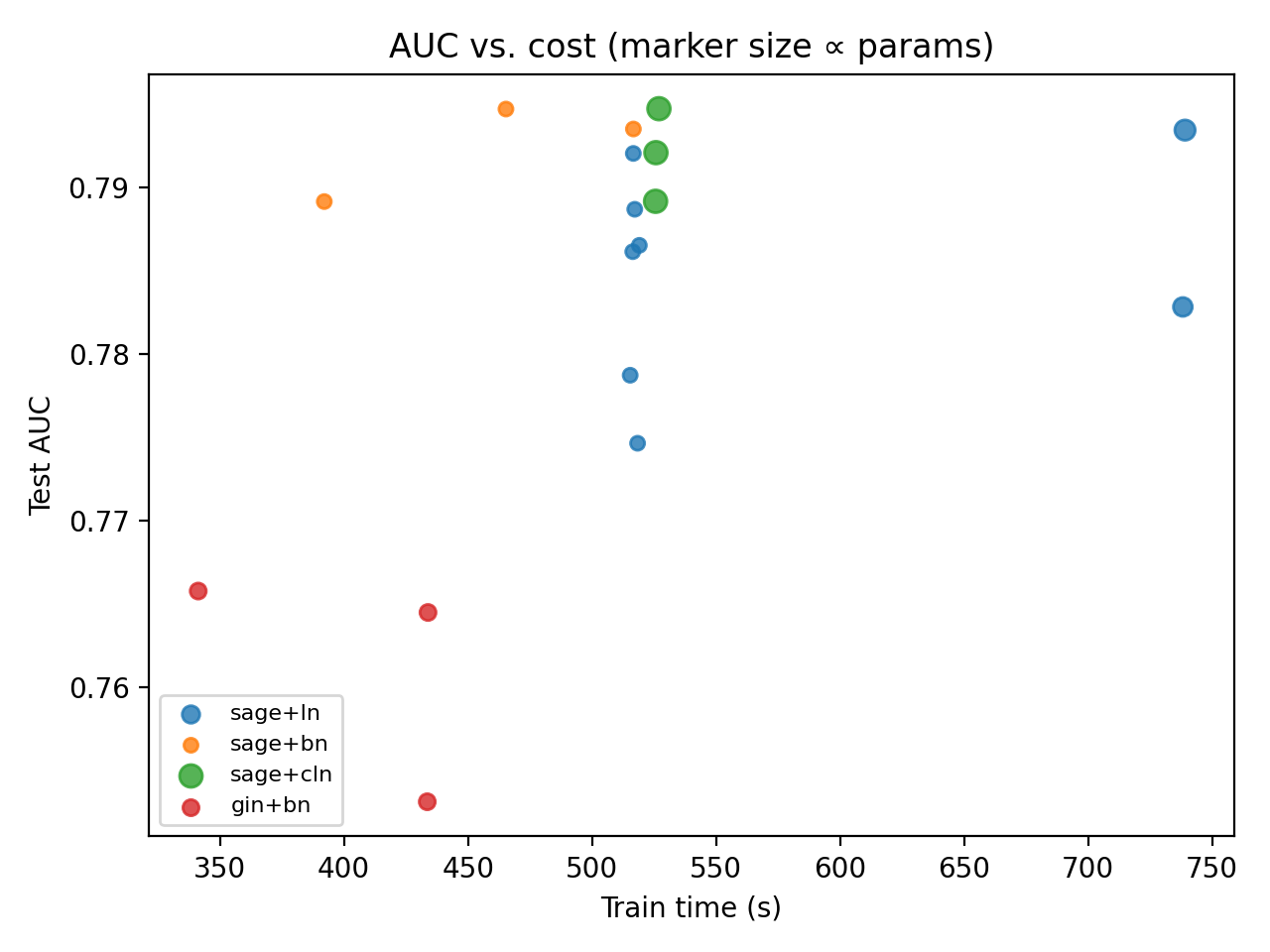}
  \caption{\textbf{AUC vs.\ training cost.} Marker size $\propto$ parameter count; color encodes model+norm. The Pareto frontier is dominated by SAGE variants.}
  \label{fig:auc-vs-cost}
\end{figure}

\paragraph{Efficiency.}
On our machine, SAGE baselines form a favorable frontier of AUC vs.\ wall-clock and VRAM. CLN adds parameters versus LN/BN but remains efficient; GIN (not shown in the main table) remains a weaker baseline even when strengthened.

\subsection{Calibration Analysis}
\label{sec:posthoc}
We apply per-label temperature scaling on validation logits and derive per-label thresholds (F$_\beta{=}1$). This markedly improves decision quality (micro-F1) and calibration (ECE) while leaving AUC essentially unchanged. Adding label-correlation smoothing (conditional-centered graph, $\lambda{=}0.1$, logit space) provides small, consistent gains.

\begin{table}[H]
\centering
\small
\begin{tabular}{lccc|ccc}
\toprule
 & \multicolumn{3}{c}{Calib (per-label T + thresholds)} & \multicolumn{3}{c}{+ Smoothing ($\lambda{=}0.1$)}\\
\cmidrule(lr){2-4}\cmidrule(lr){5-7}
\textbf{Model} & \textbf{AUC} & \textbf{F1$_\text{micro}$} & \textbf{ECE} & \textbf{AUC} & \textbf{F1$_\text{micro}$} & \textbf{ECE}\\
\midrule
SAGE+LN (sum)         & 0.792 & 0.795 & 0.188 & 0.794 & 0.796 & 0.178 \\
SAGE+CLN(desc) (sum)  & 0.795 & 0.786 & 0.183 & 0.796 & 0.787 & 0.173 \\
SAGE+BN (sum)         & 0.795 & 0.592 & 0.350 & 0.794 & 0.587 & 0.348 \\
\bottomrule
\end{tabular}
\caption{Test metrics after per-label temperature scaling \& per-label thresholds, with/without label-correlation smoothing (conditional-centered, logit space). Values are rounded from the new runs.}
\label{tab:posthoc}
\end{table}

\paragraph{Takeaways.}
Per-label temperature + thresholds dramatically improves micro-F1 (e.g., SAGE+LN from 0.14@0.5 to $\sim$0.80) with negligible AUC change. Smoothing yields small, consistent improvements and slightly better ECE for LN/CLN. BN remains the least calibrated even after post-hoc correction.

\subsection{Label correlation}

\begin{table}[H]
\centering
\small
\begin{tabular}{lcc}
\toprule
\textbf{Correlation Metric} & \textbf{Value} & \textbf{Interpretation}\\
\midrule
Number of labels & 112 & Total protein functions\\
Sparsity & 0.89\% & Very sparse label co-occurrence\\
Mean correlation & 0.009 & Low pairwise correlations\\
Max correlation & 0.061 & Some strong functional relationships\\
Min correlation & 0.0003 & Many independent functions\\
Mean outgoing correlation & 0.999996 & Very high directional consistency\\
Mean incoming correlation & 0.999996 & Balanced directional relationships\\
\bottomrule
\end{tabular}
\caption{Label correlation statistics.}
\label{tab:label-correlation}
\end{table}

\paragraph{Key findings.}
 \textbf{Sparsity:} The 0.89\% sparsity indicates that most protein function pairs are independent, with only a small fraction showing meaningful co-occurrence. \textbf{Directional relationships:} The extremely high outgoing and incoming correlations (0.999996) suggest strong directional consistency in functional dependencies. \textbf{Biological interpretation:} The low mean correlation (0.009) with some high maximum correlations (0.061) indicates that while most protein functions are independent, there exist specific functional modules with strong interdependencies.

\section{Conclusion}
Seemingly small design choices matter on \texttt{ogbn-proteins}. Using \emph{sum} for edge$\to$node feature construction consistently improves GraphSAGE; BN delivers the best AUC, while CLN matches the AUC frontier and avoids BN’s poor fixed-threshold behavior. Post-hoc per-label temperature scaling and per-label thresholds are essential for reliable multi-label decisions, and light label-correlation smoothing adds small, consistent gains. A frozen-gate MPNN collapses as expected; we outline a practical unfrozen configuration. We standardize outputs (\texttt{args.json}, \texttt{metrics.json}, \texttt{logits\_\{train,val,test\}.npz}) and ship scripts of the runs to support transparent reruns and analysis.

\paragraph{Limitations.}
We did not include unfrozen edge-network MPNN results due to time; our configuration is provided for straightforward reproduction.

\section{Future Work}
\label{sec:limitations-future}

\paragraph{Unfrozen edge-network MPNN.}
Train the edge gate end-to-end and compare to SAGE at matched params/VRAM (3 seeds). A practical starting point is:
\texttt{--backend sparse --hid 256 --gate\_hid 64 --dropout 0.1 --epochs 120 --patience 12 --lr 2e-3 --amp 0 --norm ln/cln --alpha\_chunk 2000000 --freeze\_edge 0}
(with \texttt{--backend scatter} + chunking if memory is tight). Report AUC/F1/ECE.

\paragraph{Graph Transformers with edge channels.}
Evaluate a lightweight graph transformer that ingests the 8-D edge evidence as attention biases or per-head gates. Compare against SAGE at matched params/VRAM to see if attention helps on cross-species transfer \cite{ying2021graphormer,dwivedi2021san}.

\paragraph{Stronger calibration.}
Beyond temperature scaling, try vector scaling / classwise Platt / isotonic per label; add \emph{species-conditional} temperatures (fit on validation species, apply to test). Report micro/macro F1, ECE, and reliability plots.

\paragraph{Cost-sensitive thresholds.}
Optimize thresholds for target operating points (e.g., fixed precision 90\% or fixed recall 80\%), and show precision–recall trade-offs per label family. This reflects realistic lab-screening constraints.

\paragraph{Label graph learning.}
Replace heuristic $P$ with a learned label–label graph (e.g., from a small GNN over labels) trained on validation only \cite{mlgcn2019}. Add GO hierarchy priors and compare logit- vs.\ prob-space smoothing, including per-label $\lambda$.

\end{document}

%% file: tables/main_gnn_baselines.tex
\begin{table}[H]
\centering\small
\begin{tabular}{lccccccccc}
\toprule
Model & Norm & Edge$\to$node & Layers & Hid & Edge scalar & Val AUC & Test AUC & Test F1 & Params (M)\\
\midrule
sage & bn  & sum & 3 & 512 & sum       & $0.864$ & $0.792$ & $0.096$ & 0.650 \\
sage & cln & sum & 3 & 512 & sum       & $0.855$ & $0.792$ & $0.145$ & 1.710 \\
sage & ln  & sum & 3 & 512 & learned1d & $0.850$ & $0.787$ & $0.128$ & 0.650 \\
sage & ln  & sum & 3 & 512 & sum       & $0.855$ & $0.789$ & $0.144$ & 0.650 \\
gin  & bn  & sum & 3 & 512 & sum       & $0.845$ & $0.761$ & $0.149$ & 0.852\\

\bottomrule
\end{tabular}
\caption{Main GNN baselines on \texttt{ogbn-proteins} (3 seeds).}
\label{tab:main-gnn}
\end{table}

%% file: tables/aggr_ablation.tex
\begin{table}[H]\centering\small
\begin{tabular}{lccccccccc}
\toprule
Model & Norm & Edge$\to$node & Layers & Hid & Edge scalar & Val AUC & Test AUC & Test F1 & Params\\
\midrule
sage & ln & max & 3 & 512 & sum & 0.824 ± 0.000 & 0.779 & 0.141 & 0.650\\
sage & ln & mean & 3 & 512 & sum & 0.839 & 0.775 & 0.138 & 0.650 \\
sage & ln & sum & 3 & 512 & learned1d & 0.850 & 0.787 & 0.128 & 0.650\\
sage & ln & sum & 3 & 512 & sum & 0.855 & 0.789 & 0.144 & 0.650 \\
\bottomrule
\end{tabular}
\caption{Edge$\to$node aggregation ablation for SAGE+LN (h=512, L=3).}
\label{tab:aggr}
\end{table}